\documentclass[journal]{IEEEtran}

\usepackage{graphicx}
\usepackage{cite}
\usepackage{amsmath}
\usepackage{booktabs}
\usepackage{url}
\usepackage{hyperref}
\usepackage[margin=1in]{geometry}

\begin{document}

\title{Which India Survives Translation? Narrative Homogenisation Across Indian Oral Traditions in LLMs}

\author{
  \IEEEauthorblockN{Paarth Singh Rathore}\\
  \IEEEauthorblockA{
    \textit{Department of Computer Science and Information Systems}\\
    \textit{BITS Pilani}\\
    Pilani, Rajasthan, India\\
    paarthsinghrathore1@gmail.com
  }
}

\maketitle

\begin{abstract}
Large language models (LLMs) are trained predominantly on English-language internet text that systematically over-represents certain cultural narratives, raising the concern that models flatten the diversity of non-Western storytelling traditions into a single homogenised archetype. We present a computational study examining this phenomenon across three maximally distinct Indian regional oral and literary traditions: the Rajasthani Pabuji epic, classical Tamil Sangam poetry, and Bengali folk tales. We collected authentic reference corpora for each tradition (11, 21, and 10 passages respectively) and prompted two LLMs (Claude Sonnet and Gemini) with 54 generation requests spanning three prompt types per tradition -- generic, culturally specific, and regional-language -- each repeated across independent runs. Using Sentence-BERT embeddings and cosine similarity, we measure reference drift (how closely model outputs track their own tradition's authentic texts relative to the other two) and cross-tradition convergence (how similar outputs are to each other across traditions). We find that while LLM outputs remain closer to their own tradition's reference than to others, cross-tradition similarity is high (0.52--0.66 cosine similarity) relative to what the traditions' genuine linguistic and structural distance would predict, indicating partial homogenisation. Unexpectedly, prompting in the regional language (Hindi, Tamil, or Bengali) consistently \emph{reduced} fidelity to the authentic tradition relative to English prompting, by as much as 27 percentage points for Rajasthani and Bengali traditions. We discuss this finding against conflicting prior results on multilingual prompting and argue it reflects a difference between eliciting general cultural diversity and faithfully simulating one narrow, lesser-documented oral tradition. We position this study as a lightweight, scalable complement to recent
large-scale human-annotation studies of Indian cultural misrepresentation
in LLM-generated stories.
\end{abstract}


\section{Introduction}

Large language models (LLMs) increasingly mediate how millions of people encounter, generate, and circulate cultural narratives, from short-form creative writing to AI-assisted storytelling products. Because these models are trained predominantly on English-language internet text, the cultural narratives most heavily represented in their training data are not a neutral sample of human storytelling traditions but a skewed one, dominated by Anglo-American and broader Western sources \cite{navigli2023biases, ferrara2023should, gallegos2024bias}. A growing body of work has shown that this skew manifests not only as surface-level stereotyping -- the familiar finding that a profession defaults to a particular gender or ethnicity -- but as a deeper bias in the cultural values models express when surveyed \cite{tao2024cultural}, in the entities they associate with a given cultural context \cite{naous2024having, naous2025camellia}, and in the social stereotypes encoded in identity-attribute associations for specific national contexts such as India \cite{sahoo2024indibias}.

A narrower and less explored question is what happens specifically to \emph{narrative structure} and \emph{storytelling content} when LLMs are asked to generate fiction situated within a particular cultural or regional tradition. Recent work suggests the effect here may be especially severe: Rettberg and Wigers \cite{rettberg2025ai} generated nearly twelve thousand short stories across 236 national demonyms using a single LLM and found that, despite surface-level national markers, almost all stories converge on a single underlying plot structure -- a protagonist resolving a minor community conflict by reconnecting with tradition -- which they term \emph{narrative standardisation}, a distinct and underacknowledged form of AI bias operating at the level of plot rather than word choice. Complementing this large-scale generation study, Bhagat et al. \cite{bhagat2026tales} conducted an extensive human-annotation study (108 expert annotators across 71 Indian regions and 14 languages) and found that 88\% of LLM-generated stories about Indian cultural identities contained at least one misrepresentation, with error rates rising substantially for lower-resource Indic languages and less-documented regions. Crucially, their results suggest this is not primarily a knowledge deficit -- models answer direct factual questions about the same cultural content with roughly 77\% accuracy in English (around 60\% in Indic languages, with the strongest model reaching 86\% English / 74\% Indic) -- but a failure to reliably \emph{apply} that knowledge during open-ended narrative generation.

These two studies motivate the present work but leave an open methodological gap. Rettberg and Wigers's approach is generative and large-scale but does not isolate specific, well-documented oral or literary traditions within India, treating India as a single demonym among 236. Bhagat et al.'s approach achieves fine-grained, regionally grounded measurement but requires substantial human-annotation infrastructure (over a hundred trained native-language annotators), making it costly to extend or replicate quickly. We address this gap with a lightweight, fully computational approach: rather than relying on human judges, we measure narrative homogenisation directly via sentence embeddings, comparing LLM-generated stories against authentic reference texts from three specific, well-documented, and culturally distinct Indian oral and literary traditions.

We choose three traditions deliberately selected for maximal mutual distinctiveness: the Pabuji epic, a living oral performance tradition from rural Rajasthan centred on desert heroism and cattle protection; classical Tamil Sangam poetry, a codified pre-modern literary corpus with a distinct \emph{akam}/\emph{puram} (interior/exterior) emotional grammar; and Bengali folk tales in the tradition popularised by Lal Behari Day, characterised by village settings and morally ambiguous female protagonists. If LLMs possess genuine, differentiated knowledge of these traditions, outputs prompted for each should remain closer to that tradition's authentic corpus than to the other two. If instead outputs converge toward a generic ``Indian story'' archetype regardless of which tradition was named, this constitutes direct evidence of the narrative homogenisation described by Rettberg and Wigers, operationalised here at the sub-national, tradition-specific level that their study did not examine.

We additionally test whether prompting in the relevant regional language (Hindi, Tamil, or Bengali, as appropriate to each tradition) improves fidelity relative to English prompting. This question sits at the centre of an active and unresolved debate: Wang et al. \cite{wang2025multilingual} report that multilingual prompting substantially increases response diversity and roughly halves hallucination rates relative to English prompting with equivalent cultural cues, while Naous et al. \cite{naous2024having} and Naous et al. \cite{naous2025camellia} find that operating in-language does not reliably improve, and sometimes worsens, cultural-entity preference accuracy relative to English. Our study contributes a third, narrative-specific data point to this debate.

The remainder of this paper is organised as follows. Section II reviews related work on LLM cultural bias, narrative homogenisation, and multilingual prompting effects. Section III describes our reference corpora, prompting protocol, and embedding-based analysis methodology. Section IV presents results across four analyses: reference drift, cross-tradition convergence, prompt-type effects, and language-prompting effects. Section V discusses these findings in relation to prior work. Section VI states limitations, and Section VII concludes with directions for extending this work in future
research.

\section{Related Work}

\subsection{Bias in Large Language Models}

Bias in LLMs has been extensively documented and surveyed. Navigli et al. \cite{navigli2023biases}, Ferrara \cite{ferrara2023should}, and Gallegos et al. \cite{gallegos2024bias} each provide comprehensive treatments of how training-data selection and downstream alignment choices introduce and amplify social biases across gender, age, ethnicity, religion, and culture, and survey existing approaches to measuring and mitigating these effects. These surveys establish that cultural bias is one recognised facet of a broader, well-studied problem rather than an isolated phenomenon, and they motivate the more targeted, culture-specific measurement approaches reviewed below.

\subsection{Cultural Bias and Cultural Alignment}

Tao et al. \cite{tao2024cultural} conducted a large-scale audit of five GPT models against the World Values Survey, finding that all models, irrespective of release date, cluster culturally close to English-speaking and Protestant European countries when surveyed on values such as self-expression, secularism, and trust. They further show that \emph{cultural prompting} -- explicitly assigning the model a national identity in-prompt -- reduces this cultural distance for 71--81\% of countries with later GPT-4-family models, though the strategy can backfire for countries where the model's unprompted default already approximates that country's true values, and the residual gap after correction remains substantial.

Naous et al. \cite{naous2024having} introduce CAMeL, a benchmark of over 20,000 culturally-annotated entities and naturalistic prompts contrasting Arab and Western culture, and show that sixteen language models, including Arabic-monolingual ones, systematically prefer Western-associated entities even in explicitly Arab-contextualised prompts, with a Cultural Bias Score in the 40--65\% range against an ideal near zero. They trace part of this bias to the cultural composition of common Arabic pretraining sources, finding Arabic Wikipedia paradoxically the most Western-skewed of the corpora they examined. Naous et al. \cite{naous2025camellia} extend this entity-centric methodology to nine Asian languages and six Asian cultures (including four Indian languages: Hindi, Malayalam, Marathi, and Gujarati) in the Camellia benchmark, again finding a consistent failure to prefer culturally appropriate entities even under explicit cultural grounding, with the bias gap between Asian-language and English-language entity extraction far larger than the gap observed when the same task is conducted entirely in English.

Closer to the Indian context specifically, Sahoo et al. \cite{sahoo2024indibias} introduce IndiBias, a bilingual (English/Hindi) benchmark spanning seven individual bias axes -- including caste and region, dimensions largely absent from Western-centric benchmarks such as CrowS-Pairs -- plus three intersectional axes. Evaluating ten language models, they find bias is neither uniform across models nor consistently lower in one language over the other, and that caste- and religion-linked stereotypes are measurably present and in some cases amplified relative to general-purpose benchmarks.

\subsection{Narrative Homogenisation}

Most directly relevant to the present study, Rettberg and Wigers \cite{rettberg2025ai} generated 11,800 stories across 236 national demonyms using gpt-4o-mini and found that nearly all converge on a single underlying plot skeleton -- a protagonist reconnecting with a small community through a minor, non-violent conflict resolution -- regardless of the nationality specified in the prompt. They term this \emph{narrative standardisation}, distinguishing it from the more familiar word- or image-level \emph{representational bias}, and argue it constitutes an underacknowledged structural form of AI bias. Their analysis of the small set of Indian stories in their corpus is limited to two illustrative examples, leaving open the question this paper addresses directly: whether this standardisation persists, and to what degree, when prompts target specific, well-documented sub-national Indian oral traditions rather than the demonym ``Indian'' alone.

Bhagat et al. \cite{bhagat2026tales} take a complementary, human-annotation-driven approach with TALES, a mixed-methods study combining a community-elicited taxonomy of cultural misrepresentation types with a large-scale annotation study (2,925 annotations across 540 stories, 108 native-language expert annotators spanning 71 Indian regions and 14 languages). They find 88\% of stories contain at least one misrepresentation, with rates rising sharply for lower-resource Indic languages and less-prominent regions, and -- importantly for our study's framing -- find via a derived knowledge-question benchmark that misrepresentation is largely \emph{not} explained by models lacking the relevant facts, but by a failure to reliably apply known cultural knowledge during open-ended generation. Their finding that English-language story generation actually contains a higher density of culturally specific content than Indic-language generation (attributed to lower overall engagement with cultural specificity, not greater accuracy, in non-English output) parallels, and partially anticipates, our own finding regarding regional-language prompting.

Agarwal et al. \cite{agarwal2025ai} provide a controlled-experiment complement to these generation studies, showing that when Indian and American users receive identical AI autocomplete suggestions while writing about culturally grounded topics, Indian writing converges measurably toward American stylistic and lexical norms (cross-cultural embedding similarity rising from 0.48 to 0.54 with AI assistance), while the reverse convergence is negligible. They rule out simple explanations such as grammar correction or omission of explicit cultural nouns, suggesting the homogenisation operates at a deeper stylistic level -- a finding consistent with, and reinforcing, the structural homogenisation argument advanced in \cite{rettberg2025ai}.

\subsection{Multilingual and Regional-Language Prompting Effects}

Whether prompting in a culturally relevant language improves or worsens cultural fidelity is contested in recent literature, and our regional-language finding speaks directly to this disagreement. Wang et al. \cite{wang2025multilingual} propose multilingual prompting -- combining cultural-persona cues with prompts translated into the corresponding language -- and show across four model families that it increases response diversity substantially over English-only baselines and over established diversity techniques such as high-temperature sampling, and roughly halves hallucination rates relative to English prompts carrying equivalent cultural cues without translation. In contrast, the entity-preference studies of Naous et al. \cite{naous2024having, naous2025camellia} find that operating in the target language does not reliably improve, and in some extraction tasks worsens, cultural-entity accuracy relative to English, with the Asian-language accuracy gap shrinking substantially when the identical task is instead conducted in English. Bhagat et al.'s finding \cite{bhagat2026tales} that English-language story generation contains denser, if not more accurate, cultural content than Indic-language generation sits closer to the latter position. Our own finding -- that regional-language prompting reduces fidelity to a specific, narrow oral tradition relative to English prompting -- extends this contested picture to the case of narrative generation for sub-national, lesser-documented traditions, and we discuss possible reconciliations between these conflicting results in Section V.

\section{Methodology}

\subsection{Tradition Selection}

We selected three Indian regional storytelling traditions chosen to be maximally distinct in geography, language, literary form, and underlying moral or emotional grammar, while each remaining well-documented enough to allow construction of an authentic reference corpus in English translation.

\textbf{Rajasthani (Pabuji epic).} A living oral performance epic from rural Rajasthan, traditionally sung by Bhopa performer-priests accompanying a painted narrative scroll (\emph{phad}). Thematically centred on desert heroism, cattle protection, caste loyalty, and chivalric vow-keeping. Our reference corpus is drawn from John D. Smith's English translation of the epic, freely available through the University of Cambridge's Faculty of Asian and Middle Eastern Studies.

\textbf{Tamil (Sangam poetry).} A classical literary corpus dated approximately 300 BCE--300 CE, among the oldest surviving bodies of literature in any Dravidian language, governed by a codified \emph{akam} (interior, love-themed) and \emph{puram} (exterior, heroic/martial) thematic division. Our reference corpus draws on English translations from the \emph{Purananuru} (heroic) and \emph{Kurunthokai} (love) anthologies.

\textbf{Bengali (folk tales).} Village-set oral folk narratives, drawn here from Lal Behari Day's 1883 English-language collection \emph{Folk-Tales of Bengal}, characterised by domestic settings, morally ambiguous protagonists, and frequent female-centred narrative agency, contrasting with both the martial register of the Rajasthani corpus and the codified emotional grammar of the Tamil corpus.

After cleaning (removal of footnote markers, scholarly annotation, and non-English script while preserving full narrative content), the reference corpus comprised 11 passages for Rajasthani, 21 for Tamil, and 10 for Bengali, each passage averaging 100--300 words.

\subsection{Prompting Protocol}

For each tradition we designed three prompt types, intended to test different mechanisms by which an LLM might be cued toward authentic versus generic output:

\begin{itemize}
\item \textbf{Type 1 (Generic):} names the tradition directly without further elaboration (e.g., ``Tell me a story from the Pabuji epic, the oral tradition of Rajasthan in India.'').
\item \textbf{Type 2 (Culturally specific):} supplies tradition-specific narrative or thematic detail beyond the tradition's name (e.g., specifying a Rajput hero's vow to protect cattle, set in the Marwar desert).
\item \textbf{Type 3 (Regional language):} repeats the substance of the generic prompt, but written in the tradition's associated regional language -- Hindi for Rajasthani, Tamil for Tamil Sangam, and Bengali for the Bengali folk tradition.
\end{itemize}

Each of the resulting nine prompts (three traditions $\times$ three prompt types) was issued to two LLMs -- Claude Sonnet and Gemini -- in three independent runs each, with every run conducted in a fresh conversation to avoid cross-contamination from conversational context. This yielded $9 \times 2 \times 3 = 54$ total generated outputs. All prompts instructed the model to produce approximately 150--200 words, to keep output length comparable across conditions for embedding-based comparison.

\subsection{Embedding-Based Analysis}

All reference passages and LLM outputs were converted into dense sentence embeddings using Sentence-BERT \cite{reimers2019sentence} (the \texttt{all-mpnet-base-v2} model), with embeddings L2-normalised at encoding time so that cosine similarity is equivalent to the dot product and consistent with the distance metric used in subsequent dimensionality reduction. Reference embeddings for each tradition were averaged into a single centroid vector representing that tradition's authentic narrative ``signature.''

We conducted four analyses:

\textbf{Reference drift.} For each LLM output, we computed cosine similarity to its own tradition's reference centroid (\emph{own-similarity}) and the mean cosine similarity to the other two traditions' centroids (\emph{other-similarity}). The difference (other-similarity minus own-similarity) constitutes a drift score: a positive score indicates the output is, on average, more similar to traditions it was not prompted for than to the one it was -- direct evidence of homogenisation at the level of an individual output.

\textbf{Cross-tradition convergence.} For each LLM, we computed the mean embedding of all outputs generated for a given tradition, then measured pairwise cosine similarity between these per-tradition mean vectors across all three traditions. High similarity here indicates that, independent of any reference text, the model's own outputs for different traditions resemble each other.

\textbf{Prompt-type effect.} We compared mean own-similarity scores across the three prompt types within each tradition, to test whether culturally specific prompting improves fidelity over generic prompting.

\textbf{Language-prompting effect.} We compared mean own-similarity scores between English-language prompts (Types 1 and 2 pooled) and the regional-language prompt (Type 3) within each tradition.

For visualisation, all reference and output embeddings were jointly projected into two dimensions using UMAP with cosine distance as the metric, consistent with the distance measure used in all numerical analyses.

\section{Results}

\subsection{Reference Corpus and Output Summary}

After cleaning, reference corpora comprised 11 (Rajasthani), 21 (Tamil), and 10 (Bengali) passages. The full set of 54 LLM outputs spanned both models (Claude Sonnet, Gemini) across all nine prompt conditions with three runs each.

\subsection{Reference Drift}

Table~\ref{tab:drift} reports mean cosine similarity of LLM outputs to their own tradition's reference centroid and to the mean of the other two traditions' centroids, by tradition and model.

\begin{table}[!t]
\centering
\caption{Mean Cosine Similarity to Own vs. Other Tradition References}
\label{tab:drift}
\begin{tabular}{llccc}
\toprule
Tradition & Model & Own-sim. & Other-sim. & Drift \\
\midrule
Rajasthani & Claude  & 0.626 & 0.528 & $-$0.099 \\
Rajasthani & Gemini  & 0.659 & 0.564 & $-$0.095 \\
Tamil      & Claude  & 0.602 & 0.411 & $-$0.192 \\
Tamil      & Gemini  & 0.555 & 0.456 & $-$0.099 \\
Bengali    & Claude  & 0.552 & 0.438 & $-$0.114 \\
Bengali    & Gemini  & 0.597 & 0.454 & $-$0.143 \\
\bottomrule
\end{tabular}
\end{table}

All six tradition--model combinations show negative drift, indicating that outputs remain, on average, closer to their own tradition's authentic reference than to the other two traditions. Tamil shows the largest negative drift for Claude ($-$0.192), consistent with Tamil's comparatively larger digital and scholarly footprint relative to the other two traditions. Rajasthani shows the smallest-magnitude drift and the highest absolute own-similarity scores (0.626--0.659) of the three traditions for both models, a result we treat cautiously and return to in Section V given the recurrence of highly specific named entities (e.g., ``Marwar,'' ``Pabuji'') in both the prompts and the reference corpus.

\subsection{Cross-Tradition Convergence}

Table~\ref{tab:convergence} reports pairwise cosine similarity between each model's mean output embedding per tradition.

\begin{table}[!t]
\centering
\caption{Cross-Tradition Output Similarity by Model}
\label{tab:convergence}
\begin{tabular}{llc}
\toprule
Model & Tradition Pair & Cosine Similarity \\
\midrule
Claude & Rajasthani--Tamil   & 0.518 \\
Claude & Rajasthani--Bengali & 0.652 \\
Claude & Tamil--Bengali      & 0.660 \\
Gemini & Rajasthani--Tamil   & 0.601 \\
Gemini & Rajasthani--Bengali & 0.663 \\
Gemini & Tamil--Bengali      & 0.657 \\
\bottomrule
\end{tabular}
\end{table}

Cross-tradition similarity ranges from 0.518 to 0.663, with the Rajasthani--Bengali and Tamil--Bengali pairs consistently exceeding 0.65 for both models despite these traditions sharing no common language, geographic region, or narrative form in their authentic corpora.

\subsection{Prompt-Type Effect}

Table~\ref{tab:prompttype} reports mean own-tradition similarity by prompt type.

\begin{table}[!t]
\centering
\caption{Mean Own-Tradition Similarity by Prompt Type}
\label{tab:prompttype}
\begin{tabular}{llc}
\toprule
Tradition & Prompt Type & Own-similarity \\
\midrule
Rajasthani & Type 1 (Generic)   & 0.752 \\
Rajasthani & Type 2 (Specific)  & 0.708 \\
Rajasthani & Type 3 (Regional)  & 0.468 \\
Tamil      & Type 1 (Generic)   & 0.650 \\
Tamil      & Type 2 (Specific)  & 0.582 \\
Tamil      & Type 3 (Regional)  & 0.504 \\
Bengali    & Type 1 (Generic)   & 0.651 \\
Bengali    & Type 2 (Specific)  & 0.680 \\
Bengali    & Type 3 (Regional)  & 0.393 \\
\bottomrule
\end{tabular}
\end{table}

For Rajasthani and Tamil, the generic prompt (Type 1) outperforms the culturally specific prompt (Type 2); for Bengali this is reversed, though by a small margin (0.680 versus 0.651). In all three traditions, the regional-language prompt (Type 3) scores lowest by a substantial margin.

\subsection{Language-Prompting Effect}

Table~\ref{tab:language} isolates this regional-language effect by pooling Type 1 and Type 2 (English) prompts and comparing against Type 3 (regional language) directly.

\begin{table}[!t]
\centering
\caption{English vs. Regional-Language Prompt Fidelity}
\label{tab:language}
\begin{tabular}{lccc}
\toprule
Tradition & English & Regional & Difference \\
\midrule
Rajasthani & 0.730 & 0.468 & $-$0.262 \\
Tamil      & 0.616 & 0.504 & $-$0.113 \\
Bengali    & 0.665 & 0.393 & $-$0.273 \\
\bottomrule
\end{tabular}
\end{table}

Regional-language prompting reduced own-tradition fidelity for all three traditions, with the smallest drop for Tamil ($-$0.113) and the largest for Bengali ($-$0.273) and Rajasthani ($-$0.262).

\subsection{Embedding Space Visualisation}

\begin{figure}[!t]
\centering
\includegraphics[width=\columnwidth]{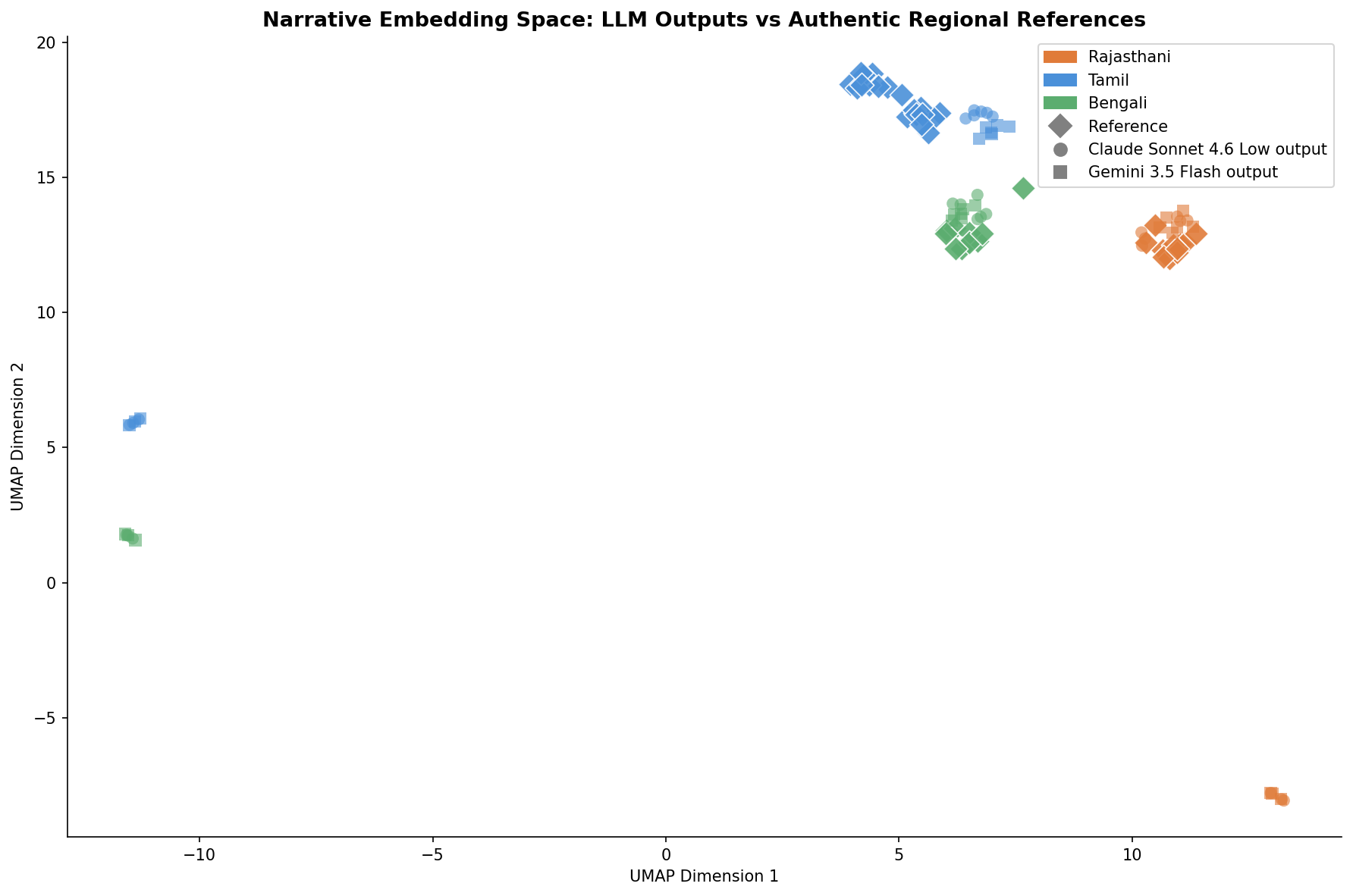}
\caption{UMAP projection (cosine metric) of Sentence-BERT embeddings for all reference passages (diamonds) and LLM outputs (circles: Claude; squares: Gemini), coloured by tradition. Distinct clusters are visible per tradition, with reference and model-output points co-locating within each cluster.}
\label{fig:umap}
\end{figure}

Fig.~\ref{fig:umap} shows a two-dimensional UMAP projection (cosine metric) of all reference passages and LLM outputs. Reference passages and outputs for each tradition form visually distinct, spatially separated clusters, with model outputs (both Claude and Gemini) co-locating near their corresponding tradition's reference cluster rather than collapsing into a single undifferentiated region.

\section{Discussion}

\subsection{Partial but Real Homogenisation}

The negative drift scores across all six tradition--model combinations (Table~\ref{tab:drift}) indicate that the two LLMs tested are not entirely indifferent to which of the three traditions they are prompted for: outputs systematically track their own tradition's authentic reference more closely than the other two. This places our finding in partial tension with the strongest reading of Rettberg and Wigers's narrative-standardisation thesis \cite{rettberg2025ai}, which might predict near-zero differentiation across prompted identities. However, the cross-tradition convergence results (Table~\ref{tab:convergence}) complicate this picture considerably: similarity scores of 0.65 or higher between traditions as structurally and linguistically distinct as Rajasthani epic and Bengali folk tale are difficult to reconcile with genuine, differentiated cultural knowledge, particularly given that all three reference corpora are themselves well-separated in the same embedding space (Fig.~\ref{fig:umap}). We read this as evidence for a model that possesses some genuine tradition-specific signal -- sufficient to keep own-similarity above other-similarity -- layered on top of a strong shared narrative substrate that pulls all three traditions' outputs toward one another. This is broadly consistent with Bhagat et al.'s knowledge-versus-application distinction \cite{bhagat2026tales}: the models may ``know'' enough about each tradition to avoid total collapse, while still defaulting, at the level of narrative execution, to shared structural and stylistic patterns.

\subsection{The Regional-Language Finding in Context}

Our clearest and most surprising result is that prompting in the regional language associated with each tradition consistently reduced fidelity to that tradition's authentic corpus, with drops as large as 27 percentage points (Table~\ref{tab:language}). This result sits awkwardly alongside Wang et al.'s finding \cite{wang2025multilingual} that multilingual prompting roughly halves hallucination rates and substantially increases cultural diversity relative to English prompting with equivalent cultural cues. We propose this divergence is best explained by a difference in task goals rather than a direct contradiction. Wang et al.'s diversity metric rewards a \emph{broad}, varied sampling across many cultures and languages simultaneously; our fidelity metric instead asks whether output for \emph{one specific, narrow} tradition stays close to that tradition's own authentic textual record. A model may successfully access broader, more varied cultural associations when prompted multilingually -- exactly as Wang et al. find -- while simultaneously failing to narrow in on the specific stylistic and narrative conventions of one particular under-documented oral tradition such as Pabuji, for which regional-language training data is almost certainly far sparser than general-purpose Hindi text. This account is consistent with the entity-level findings of Naous et al. \cite{naous2024having, naous2025camellia}, who likewise find that operating in a target language does not reliably improve, and sometimes worsens, fine-grained cultural accuracy relative to English, and with Bhagat et al.'s observation \cite{bhagat2026tales} that English-language story generation contains denser culturally specific content than Indic-language generation for the same prompts. Our comparatively smaller language-effect drop for Tamil (-0.113, versus -0.262 and -0.273 for Rajasthani and Bengali) is consistent with this resource-based account: Tamil has substantially greater digital and scholarly representation than either the Pabuji oral tradition or the specific Bengali folk corpus used here, and a similar resource-stratification pattern is reported directly by Bhagat et al. across their language tiers.

A further mechanistic possibility, suggested by Rettberg and Wigers's own incidental finding regarding prompt brittleness in non-English generation \cite{rettberg2025ai}, is that regional-language prompts may shift not only the cultural register of the response but its underlying register or genre entirely -- for instance, eliciting a more devotional or formally Sanskritised register in Hindi that is itself stylistically distant from the oral-epic register of our Pabuji reference corpus, independent of any cultural knowledge gap per se. Disentangling these accounts is not possible with the present study design
and is a priority for future work, described in Section VII.

\subsection{The Anomalous Rajasthani Result}

Rajasthani shows both the smallest-magnitude drift and the highest absolute own-similarity of the three traditions (Table~\ref{tab:drift}), despite the Pabuji epic being, by most measures, the least digitally documented of the three traditions studied. We treat this result cautiously rather than as evidence of unexpectedly strong model knowledge of Pabuji specifically. Both our Type 1 and Type 2 English prompts for this tradition explicitly named distinctive entities present in the reference corpus itself (``Pabuji,'' ``Marwar,'' cattle-protection vows), and it is plausible that high own-similarity scores partly reflect superficial lexical overlap on these named entities rather than deeper structural or stylistic fidelity to the oral-epic form. This concern is reinforced by the prompt-type results (Table~\ref{tab:prompttype}), where the more elaborately cued Type 2 prompt for Rajasthani \emph{underperforms} the bare Type 1 prompt -- the opposite of what genuine cultural knowledge being more fully activated by richer cuing would predict, and more consistent with an account in which specific named-entity overlap, rather than narrative fidelity, is driving the own-similarity signal for this tradition. We flag this explicitly as a methodological limitation requiring resolution in future work, discussed further below.

\section{Limitations}

This study has several limitations that bound the strength of its
conclusions and motivate its extension into a larger study.

First, only two LLMs were evaluated (Claude Sonnet, Gemini); GPT-family models were excluded due to access constraints, limiting the generalisability of our findings across the broader LLM landscape.

Second, our reference corpora, while authentic, are themselves English translations of oral or classical-language originals (Hindi/Marwari for Pabuji, classical Tamil for Sangam poetry, Bengali for the folk tales). Comparing English-language model outputs against English \emph{translations} of these traditions, while comparing regional-language model outputs against the same English references via embedding models trained primarily on English and high-resource-language data, plausibly inflates the regional-language fidelity penalty we report in Section IV-D, since translation-mediated comparison is not equivalent for the two conditions. This is among the most important confounds to resolve in extending this work, for instance via native-language reference corpora and multilingual embedding models.

Third, our use of sentence-embedding cosine similarity as a homogenisation proxy is a lightweight, scalable substitute for, not a replacement of, fine-grained human judgment of the kind employed at much greater scale by Bhagat et al. \cite{bhagat2026tales}. Embedding similarity captures broad semantic and topical overlap but cannot distinguish, for instance, superficial named-entity matching from genuine structural or stylistic fidelity to a narrative tradition -- a distinction directly relevant to our discussion of the Rajasthani result in Section V-C.

Fourth, the reference corpora themselves are of uneven size (11, 21, and 10 passages for Rajasthani, Tamil, and Bengali respectively), reflecting genuine differences in the availability of freely accessible, high-quality English translations across these three traditions rather than a deliberate sampling choice; this asymmetry may itself partially confound cross-tradition comparison.

Finally, with 54 total outputs across three runs per condition, this study is explicitly scoped as an initial investigation rather than a
comprehensive evaluation, and statistical significance testing of the differences reported in Section IV was not conducted given the modest per-condition sample sizes.

\section{Conclusion and Future Work}

This study finds that two widely used LLMs, when prompted to generate stories from three maximally distinct Indian regional oral and literary traditions, produce outputs that remain measurably closer to their own tradition's authentic reference corpus than to the other two traditions, yet still show high (0.52--0.66) cross-tradition output similarity relative to what these traditions' genuine cultural and linguistic distance would predict -- evidence for partial, rather than total, narrative homogenisation. We further find that prompting in the regional language associated with each tradition consistently reduces, rather than improves, fidelity to that tradition's authentic narrative register, a result that adds a new, narrative-specific data point to an unresolved debate in the multilingual-prompting literature.

This work extends recent large-scale investigations of narrative archetypes
and cultural bias in LLM-generated text \cite{rettberg2025ai} to the specific
case of Indian regional and sub-national oral traditions. Planned extensions include: expanding the reference corpus to additional traditions and a larger number of passages per tradition; incorporating additional LLMs, including open-weight models, to test whether homogenisation patterns are consistent across model families and training regimes; constructing native-language (rather than English-translated) reference corpora paired with multilingual embedding models to disentangle genuine cultural-fidelity effects from translation-mediated artefacts; combining the present embedding-based approach with structured human annotation, following the TALES taxonomy of misrepresentation types \cite{bhagat2026tales}, to validate whether high embedding similarity to a tradition's reference corpus corresponds to genuine narrative and structural fidelity rather than superficial lexical overlap, directly addressing the concern raised in Section V-C regarding the Rajasthani results; and systematically varying prompt specificity and persona-cuing strategies, following the multilingual and multicultural prompting framework of Wang et al. \cite{wang2025multilingual}, to test whether the language-prompting penalty observed here is specific to narrow, lesser-documented oral traditions or generalises across resource levels.

\bibliographystyle{IEEEtran}
\bibliography{references}

@inproceedings{naous2024having,
  title={Having beer after prayer? measuring cultural bias in large language models},
  author={Naous, Tarek and Ryan, Michael J and Ritter, Alan and Xu, Wei},
  booktitle={Proceedings of the 62nd annual meeting of the association for computational linguistics (volume 1: Long papers)},
  pages={16366--16393},
  year={2024}
}

@article{naous2025camellia,
  title={Camellia: Benchmarking cultural biases in llms for asian languages},
  author={Naous, Tarek and Savit, Anagha and Catalan, Carlos Rafael and Guo, Geyang and Lee, Jaehyeok and Lee, Kyungdon and Dizon, Lheane Marie and Ye, Mengyu and Kothari, Neel and Singh, Sahajpreet and others},
  journal={arXiv preprint arXiv:2510.05291},
  year={2025}
}

@inproceedings{sahoo2024indibias,
  title={IndiBias: A benchmark dataset to measure social biases in language models for Indian context},
  author={Sahoo, Nihar and Kulkarni, Pranamya and Ahmad, Arif and Goyal, Tanu and Asad, Narjis and Garimella, Aparna and Bhattacharyya, Pushpak},
  booktitle={Proceedings of the 2024 Conference of the North American Chapter of the Association for Computational Linguistics: Human Language Technologies (Volume 1: Long Papers)},
  pages={8786--8806},
  year={2024}
}

@inproceedings{agarwal2025ai,
  title={Ai suggestions homogenize writing toward western styles and diminish cultural nuances},
  author={Agarwal, Dhruv and Naaman, Mor and Vashistha, Aditya},
  booktitle={Proceedings of the 2025 CHI conference on human factors in computing systems},
  pages={1--21},
  year={2025}
}

@article{tao2024cultural,
  title={Cultural bias and cultural alignment of large language models},
  author={Tao, Yan and Viberg, Olga and Baker, Ryan S and Kizilcec, Ren{\'e} F},
  journal={PNAS nexus},
  volume={3},
  number={9},
  pages={pgae346},
  year={2024},
  publisher={Oxford University Press US}
}

@inproceedings{wang2025multilingual,
  title={Multilingual prompting for improving llm generation diversity},
  author={Wang, Qihan and Pan, Shidong and Linzen, Tal and Black, Emily},
  booktitle={Proceedings of the 2025 Conference on Empirical Methods in Natural Language Processing},
  pages={6378--6400},
  year={2025}
}

@inproceedings{reimers2019sentence,
  title={Sentence-bert: Sentence embeddings using siamese bert-networks},
  author={Reimers, Nils and Gurevych, Iryna},
  booktitle={Proceedings of the 2019 conference on empirical methods in natural language processing and the 9th international joint conference on natural language processing (EMNLP-IJCNLP)},
  pages={3982--3992},
  year={2019}
}

@article{rettberg2025ai,
  title={AI-generated stories favour stability over change: homogeneity and cultural stereotyping in narratives generated by gpt-4o-mini},
  author={Rettberg, Jill Walker and Wigers, Hermann},
  journal={arXiv preprint arXiv:2507.22445},
  year={2025}
}

@inproceedings{bhagat2026tales,
  title={Tales: A taxonomy and analysis of cultural representations in llm-generated stories},
  author={Bhagat, Kirti and Bhatt, Shaily and Velagapudi, Athul and Vashistha, Aditya and Dave, Shachi and Pruthi, Danish},
  booktitle={Proceedings of the 2026 CHI Conference on Human Factors in Computing Systems},
  pages={1--26},
  year={2026}
}

@article{navigli2023biases,
  title={Biases in large language models: origins, inventory, and discussion},
  author={Navigli, Roberto and Conia, Simone and Ross, Bj{\"o}rn},
  journal={ACM Journal of Data and Information Quality},
  volume={15},
  number={2},
  pages={1--21},
  year={2023},
  publisher={ACM New York, NY}
}

@article{ferrara2023should,
  title={Should chatgpt be biased? challenges and risks of bias in large language models},
  author={Ferrara, Emilio},
  journal={arXiv preprint arXiv:2304.03738},
  year={2023}
}

@article{gallegos2024bias,
  title={Bias and fairness in large language models: A survey},
  author={Gallegos, Isabel O and Rossi, Ryan A and Barrow, Joe and Tanjim, Md Mehrab and Kim, Sungchul and Dernoncourt, Franck and Yu, Tong and Zhang, Ruiyi and Ahmed, Nesreen K},
  journal={Computational linguistics},
  volume={50},
  number={3},
  pages={1097--1179},
  year={2024},
  publisher={MIT Press}
}

\end{document}